\documentclass[10pt,twocolumn,letterpaper]{article}

\usepackage{cvpr}              

\usepackage{booktabs}
\usepackage{pifont}
\newcommand{\xmark}{\ding{55}}%

\definecolor{cvprblue}{rgb}{0.21,0.49,0.74}
\usepackage[pagebackref,breaklinks,colorlinks,allcolors=cvprblue]{hyperref}

\def\paperID{29461} 
\def\confName{CVPR}
\def\confYear{2026}

\title{CropVLM: Learning to Zoom for Fine-Grained Vision-Language Perception}

\author{
Miguel Carvalho \quad
Hélder Dias \quad
Bruno Martins \\
INESC-ID, Instituto Superior Técnico, University of Lisbon \\
{\tt\small \{miguelcarvalho00, helder.dias, bruno.g.martins\}@tecnico.ulisboa.pt}
}

\begin{document}
\maketitle
\begin{abstract}
Vision-Language Models (VLMs) often struggle with tasks that require fine-grained image understanding, such as scene-text recognition or document analysis, due to perception limitations and visual fragmentation. To address these challenges, we introduce CropVLM as an external low-cost method for boosting performance,  enabling VLMs to dynamically ''zoom in'' on relevant image regions, enhancing their ability to capture fine details. CropVLM is trained using reinforcement learning, without using human-labeled bounding boxes as a supervision signal, and without expensive synthetic evaluations. The model is trained once and can be paired with both open-source and proprietary VLMs to improve their performance. Our approach delivers significant improvements on tasks that require high-resolution image understanding, notably for benchmarks that are out-of-domain for the target VLM, without modifying or fine-tuning the VLM, thus avoiding catastrophic forgetting.
\end{abstract}
    
\section{Introduction}
\label{sec:intro}

Recent VLMs have demonstrated impressive capabilities in understanding and reasoning over visual contents~\cite{alayrac2022flamingo,liu2023visual,chen2024internvl}. Still, despite their success, these models face significant limitations when confronted with tasks that require fine-grained visual perception, such as document analysis, scene-text recognition, or detailed object identification. A primary constraint lies in the input resolution, as most mainstream VLMs leverage pre-trained vision encoders that can take as input images at relatively low resolutions -- e.g., 224×224 \cite{radford2021learning} or 336×336 pixels \cite{liu2023visual} -- causing crucial fine details to become indiscernible.

\begin{figure}[htbp]
    \centering
    \includegraphics[width=\columnwidth]{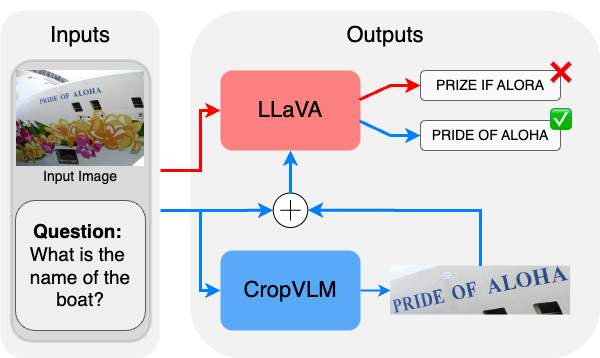}
    \caption{Overview of \textbf{CropVLM} paired with LLaVA. \textbf{CropVLM} dynamically selects informative image regions to boost fine-grained perception while keeping the target VLM frozen.}
    \label{fig:cropvlm}
\end{figure}

For instance, the LLaVA-1.5 \cite{liu2023visual} model, with an input resolution of 336x336 pixels, is effective for general visual reasoning, but it struggles significantly when processing small text or when performing detailed visual analysis \cite{liu2024ocrbench}. 
A straightforward solution involves increasing the input resolution uniformly. However, this introduces a prohibitive computational burden with VLMs based on the Transformer architecture. Recent work by \citet{shi2025scaling} or \citet{cai2024matryoshka} demonstrates that even advanced models use only a small number of image tokens to answer most requests, suggesting that uniform high-resolution processing is inefficient and unnecessary. Alternative approaches have attempted to address these limitations through architectural modifications \cite{cai2024matryoshka, shi2024we} or specialized fine-tuning \cite{shao2024visual}. While effective in controlled settings, these methods often require extensive model retraining, risking catastrophic forgetting and poor generalization to out-of-domain scenarios. Furthermore, these strategies are typically not applicable to proprietary models, given their closed weights.

\begin{table*}[ht]
\centering
\scalebox{1.0}{ 
\begin{tabular}{lcccc}
\toprule
\textbf{Method} & \textbf{Region Selection} & \textbf{Annotations} & \textbf{Optimization} & \textbf{\# of Crops}\\
\midrule
SEAL \cite{wu2024v} & LLM-Guided Search & \xmark &  LLM-Only & Multiple \\
SEMCLIP \cite{li2025semantic} & Semantic & \xmark & Embedding-Guided & Single\\
ViCrop \cite{zhang2025mllms} & Semantic & \xmark & Attention-Heuristic & Single \\
Visual-CoT \cite{shao2024visual} & Manual &  Human \& Synthetic & SFT & Single \\
VisRL \cite{chen2025visrl} & Preference &  Model Validated & DPO & Single \\ 
UV-CoT \cite{zhao2025unsupervised} & Preference &  Model Validated & DPO & Single \\ 
Visual-RFT \cite{liu2025visual} & Preference & Human & GRPO & Single\\
DeepEyes \cite{zheng2025deepeyes} & Preference & \xmark & GRPO & Multiple\\
Chain-of-Focus \cite{zhang2025chain} & Preference & \xmark & GRPO & Multiple\\
Mini-o3 \cite{lai2025mini} & Preference & \xmark & GRPO & Multiple\\
\midrule
\textbf{CropVLM (Ours)} & Preference & \xmark & GRPO & Single\\
\bottomrule
\end{tabular}
}
\caption{Comparison of visual grounding methods used together with VLMs. CropVLM supports both open- and closed-weight settings, and it does not require the use of direct supervision with human-labeled or synthetic bounding boxes.}
\label{tab:related-work-comparison}
\end{table*}

The fundamental challenge lies in the inability of current VLMs to dynamically adjust the visual focus across different spatial regions based on the task at hand, even when guided by detailed textual prompts \cite{zhao2025unsupervised}. This limitation highlights the critical need for a flexible approach that can adaptively allocate computational resources to the most relevant parts of an input image. 

A computationally efficient alternative involves adaptive selection, where a high-resolution image is decomposed into a global low-resolution view and one or more high-resolution crops of salient regions. This approach offers several advantages: it reduces computational overhead by processing a low-resolution overview alongside small and focused crops, avoiding the quadratic cost of full high-resolution encoding while preserving fine-grained details where needed; it improves task-aware efficiency by selectively encoding only relevant regions, which avoids unnecessary computation on less informative areas, and enhances inference speed and memory usage; and it ensures scalability even as input resolutions increase, since the crop selection mechanism (whether learned or heuristic-based) can dynamically adjust to particular computational constraints.

Targeting the aforementioned advantages, this paper introduces \textbf{CropVLM}\footnote{The code is made available at \url{https://github.com/miguelscarv/cropvlm}} as a reinforcement learning-based approach that enables VLMs to dynamically "zoom in" on relevant image regions, without requiring ground-truth bounding boxes as the supervision signal, and without a separate evaluator model guiding the training. Our method enhances existing VLMs with a lightweight cropping network (i.e., with only 256M parameters), which identifies task-relevant regions for finer-detail image processing. We leverage reinforcement learning to train the region selection model, eliminating the need for expensive human annotations in the form of bounding boxes. Our approach functions as a modular component that can be paired with both open-source and proprietary VLMs, without requiring modifications over the target model.

\textbf{CropVLM}, which is illustrated in Figure \ref{fig:cropvlm}, significantly improves performance in tasks that require fine-grained visual understanding, particularly for high-resolution images that are outside the training domain, and without fine-tuning the target VLM, thus avoiding catastrophic forgetting. Furthermore, by focusing computational resources on the most informative image regions, \textbf{CropVLM} achieves the benefits of high-resolution processing without incurring the full computational cost.

\section{Related Work}
\label{sec:related}

Despite their impressive capabilities, VLMs exhibit notable limitations in visual perception, particularly in tasks requiring fine-grained understanding \citep{tong2024eyes}. These challenges are especially pronounced in tasks such as document understanding, where precise visual parsing is critical. While VLMs often excel at textual reasoning, their ability to interpret fine grained details, e.g. such as small text regions \citep{liu2024ocrbench} or small objects, remains suboptimal, motivating recent efforts to improve visual grounding.

\paragraph{Improving Visual Perception.} 
A common approach to improving visual perception is Supervised Fine-Tuning (SFT) on task-specific datasets. Alternatively, other approaches focus on architectural improvements. For example, Matryoshka (M3) \citep{cai2024matryoshka} uses progressive training to enhance high-resolution reasoning, and S2 \citep{shi2024we} adapts models across visual granularities. The PS3 approach \citep{shi2025scaling} scales pre-training to 4K resolution by contrasting local image regions with detailed captions. Region-aware strategies such as Visual-CoT \citep{shao2024visual} use bounding-box supervision to help models focus on task-relevant regions, but often rely on expensive annotations or narrow training distributions.

\paragraph{Intention-Driven Visual Models.} 
To make visual grounding more adaptive, intention-driven methods dynamically prioritize image regions based on user queries. These methods fall into three broad categories: (1) \textit{LLM-guided querying}, such as SEAL \citep{wu2024v}, SEMCLIP \citep{li2025semantic}, or SoM-LLaVA \citep{yan2024list}, which guide region selection using search, textual semantics, or visual tags; (2) \textit{cropping-based preprocessing}, as seen in Visual-CoT \citep{shao2024visual} and ViCrop \citep{zhang2025mllms}, which crop inputs before answering the request; and (3) \textit{preference-based refinement}, such as VisRL \citep{chen2025visrl}, which optimize cropping through reward models or preference feedback. While these techniques improve relevance and efficiency, they often require expensive human or synthetic annotations. In contrast, the proposed  \textbf{CropVLM} method simplifies the process of learning how to focus without requiring explicit region supervision.

\paragraph{Preference Learning in VLMs.}
Preference optimization has emerged as a scalable framework for aligning models with human intent \citep{ouyang2022training}. Visual-CoT \citep{shao2024visual} applies supervised fine-tuning using annotated regions, while UV-CoT \citep{zhao2025unsupervised} and  VisRL \citep{chen2025visrl} remove this bottleneck by generating synthetic region preferences for Direct Preference Optimization (DPO) via textual feedback from general VLMs.
The Group Relative Policy Optimization (GRPO) \citep{shao2024deepseekmath} approach offers advantages over DPO in simplified reward modeling and stability. For example, Visual-RFT \citep{liu2025visual} applies GRPO to visual perception and image classification tasks, using verifiable rewards based on ground-truth bounding boxes and labels. Recent studies such as DeepEyes~\citep{zheng2025deepeyes}, Chain-of-Focus~\citep{zhang2025chain}, or Mini-o3~\citep{lai2025mini} extend this line of research through multi-turn region-of-interest reasoning. While powerful, these approaches require multiple reasoning steps at inference, leading to substantially higher test-time cost. Moreover, it remains unclear how to identify the most informative crop within a model’s reasoning trace, to effectively assist a target VLM in answering a query. Our work instead applies GRPO to single-pass visual cropping, achieving alignment without manual or synthetic bounding boxes, and enabling scalable optimization, even for closed-weight or out-of-domain settings.


\paragraph{Summary.} A comparative summary of recently proposed approaches for improving the visual comprehension abilities of VLMs is shown in Table~\ref{tab:related-work-comparison}.
\section{Methods}
\label{sec:methods}

\begin{figure*}[htbp]
    \centering
    \includegraphics[width=0.8\textwidth]{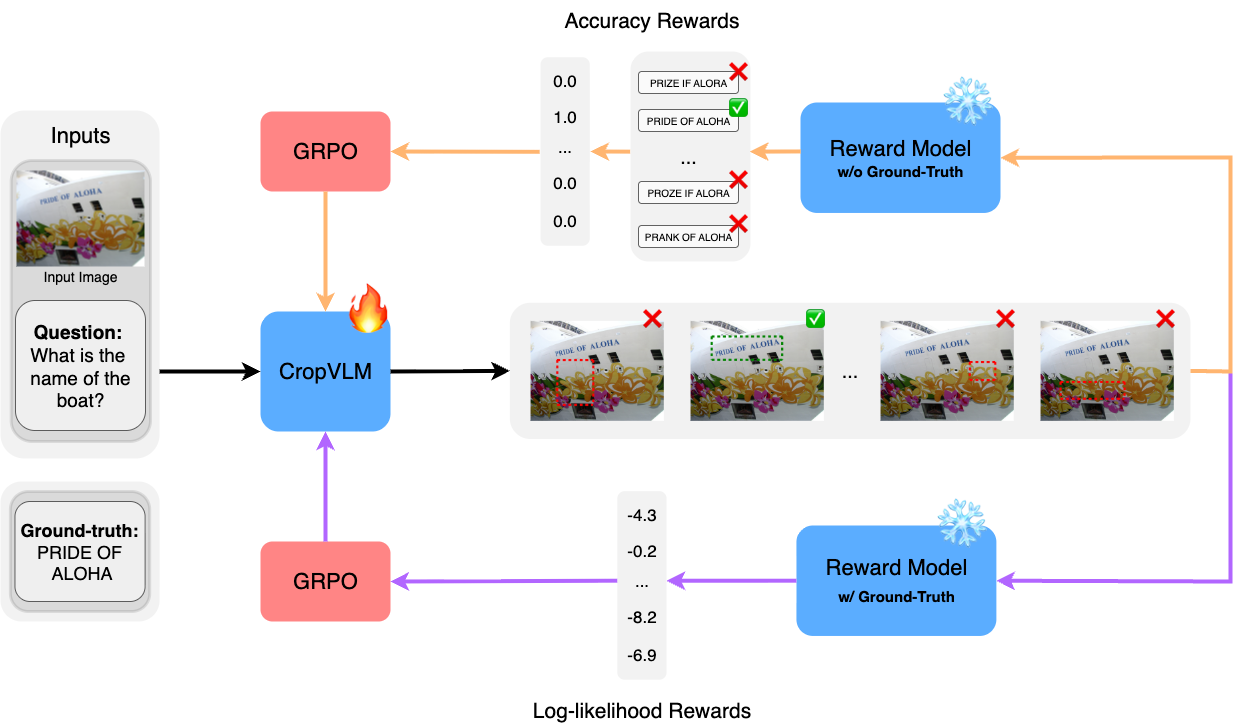}
    \caption{The overall CropVLM training procedure. The orange and purple lines represent training with an accuracy-based reward or with a log-likelihood reward, respectively.}
    \label{fig:cropvlmtraining}
\end{figure*}

This section introduces the \textbf{CropVLM} training method. In brief, \textbf{CropVLM} is a relatively small Vision-Language Model (VLM) trained to produce a single bounding box per request-image pair, aiming to improve downstream performance with a target (separate) VLM. The bounding box is generated as text in the format $[x_1, y_1, x_2, y_2]$, where $(x_1, y_1)$ denotes the top-left corner and $(x_2, y_2)$ the bottom-right corner of the crop, and where each coordinate corresponds to a percentage of the width and height of the image, so as to generalize to differently sized images.

We initialize \textbf{CropVLM} with a model already capable of generating bounding box-like outputs, albeit not very informative ones. The key idea is to train this model such that it learns to generate useful crops through reinforcement learning, without using ground-truth bounding boxes, as datasets with such labels are limited and human annotations are often suboptimal for this task -- see Appendix \ref{appendix:expanding-tight} in the supplementary material.

\paragraph{GRPO.} The specific reinforcement learning algorithm employed in our training procedure is Group Relative Policy Optimization (GRPO) \citep{shao2024deepseekmath}, i.e. a variant of Proximal Policy Optimization (PPO) \citep{schulman2017proximal} which simplifies training by avoiding the need for a value model. This algorithm allows us to optimize the bounding box generation policy in an unsupervised manner. Unlike other reinforcement learning methods such as PPO, that require a separate critic model to evaluate policy performance, GRPO directly compares groups of candidate responses. For a given input query $q$, the algorithm generates $G$ distinct responses $\{o_1, o_2, \ldots, o_G\}$ from the current policy $\pi_{\theta_{\text{old}}}$. It then evaluates these responses and assigns rewards $\{r_1, r_2, \ldots, r_G\}$ based on their quality. GRPO normalizes the rewards by computing their mean and standard deviation, this way determining the relative quality $A_i$ of each response:

\begin{equation}
A_i = \frac{r_i - \text{mean}(\{r_1, \ldots, r_G\})}{\text{std}(\{r_1, \ldots, r_G\})}.
\end{equation}

The normalization process enables the model to identify and favor higher-quality responses within each group. By training the policy to generate outputs similar to those that received higher relative rewards, GRPO effectively improves performance without requiring supervised fine-tuning or an additional value function approximator, making the training process more efficient and straightforward.

\paragraph{Reward Design.} Previous work has shown that concatenating the full image with a relevant image crop can improve VQA accuracy \cite{zhang2025mllms}. We apply a similar strategy by combining the full image and the crop, afterwards passing these inputs to a target VLM from which a reward signal is computed, as shown in Figure \ref{fig:cropvlmtraining}.

We experimented with two reward formulations:
\begin{itemize}
    \item \textbf{Accuracy-Based Reward:} The reward VLM answers requests using two images (i.e., the original plus the crop) as input. The reward is computed as the accuracy between the model's prediction and the ground-truth answers. The specific accuracy metrics that were used are described in Subsection \ref{sec:setup}.
    
    \item \textbf{Likelihood-Based Reward:} The most common ground-truth answer for the request is extracted from the dataset. The VLM then computes the log-likelihood of this answer given the corresponding input. This formulation allows for faster, single-pass reward estimation, without requiring autoregressive generation. Formally, we define the likelihood-based reward as follows:
    \begin{equation}
    \small \hspace{-0.5cm}
        \mathrm{R}(I_o, I_c, q, a^*) = \sum_{t=1}^{T} \log p_{\theta}(a^*_t | I_o, I_c, q, a^*_{<t}),
    \end{equation}
    where $p_{\theta}(a^*_t | I_o, I_c, q, a^*_{<t})$ is the probability assigned by the reward VLM to the $t$-th token of the ground-truth answer $a^*$, given the preceding tokens $a^*_{<t}$, the original image $I_o$, the cropped image $I_c$, and the request $q$.
\end{itemize}

\paragraph{Comparison to Alternative Methods.}
Our reward formulations offer key advantages over existing approaches. Unlike VisRL \citep{chen2025visrl}, which uses general-purpose VLMs for textual rewards without specialized training, our method directly measures performance using quantifiable metrics aligned with the downstream task. Additionally, our GRPO-based approach achieves higher data efficiency by utilizing all $G$ sampled bounding boxes during training, rather than retaining only two responses per query, as in DPO-based methods \citep{zhao2025unsupervised}. By grounding rewards in objective measures rather than model-generated judgments, our approach also reduces reward hacking and ensures that the improvements indeed correlate with genuine performance gains.
\section{Experiments}
\label{sec:experiments}

This section details the datasets and data transformations that were used to train \textbf{CropVLM}. We provide detailed hyperparameter choices and discuss the obtained results.

\subsection{Experimental Setup}
\label{sec:setup}
For all subsequent experiments, we select the 256M SmolVLM Instruct model \citep{marafioti2025smolvlm} to create \textbf{CropVLM} cropping networks, as this model can process images with varying resolutions and is small in size.

\paragraph{Datasets.} Fine-grained high-resolution image understanding is essential for tasks such as scene-text recognition and document understanding. These tasks often involve images with only a single and relatively small zone of interest per query. For these reasons, we selected the following training datasets: TextVQA \citep{singh2019towards} and ST-VQA \citep{biten2019scene} for scene-text understanding, DocVQA \citep{mathew2021docvqa} for document analysis, and InfographicsVQA \citep{mathew2022infographicvqa} for understanding graphs and infographics. This mixture results in a total of 124k image-question pairs. Since SmolVLM is not originally capable of generating valid bounding boxes in the format we described in Section \ref{sec:methods}, we randomly split the above training mixture into 2 equally-sized datasets, where one is used to create a synthetic seed bounding box dataset to finetune SmolVLM on, using an initial step of Supervised Fine-Tuning (SFT), and the other is used to do GRPO, after the aforementioned SFT step. To create the seed bounding box dataset we prompted the Qwen 2.5-VL 7B Instruct \citep{bai2025qwen2} model with the instruction "\textit{Outline the region in the image that would help answer the following question: \{QUESTION\} Output the coordinates in JSON format.}", to obtain a synthetic bounding box dataset. Our goal is to give the model the basic ability of generating bounding boxes in the intended format. Since the Qwen 2.5-VL model generates absolute coordinate bounding boxes, an extra normalization step is performed to comply with the previously designed format. 
Additionally, upon inspection of the generated bounding boxes, we observed that the ones with smaller relative areas tend to not include the correct area of interest within the image. To address this issue, we expand the generated bounding boxes according to their relative area percentiles, using the expansion factors reported in Table~\ref{tab:bbox-expansion}.

\begin{table}[ht]
\centering
\caption{Bounding box expansion factors based on relative area percentiles of bounding boxes to full image size.}
\label{tab:bbox-expansion}
\begin{tabular}{ccc}
\toprule
\textbf{Relative Area (\%)} & \textbf{Percentile} & \textbf{Expansion Factor} \\
\midrule
$<$ 0.16\% & Below 20th & 45x \\
0.16\% - 0.38\% & 20th - 40th & 10x \\
0.38\% - 0.91\% & 40th - 60th & 4x \\
0.91\% - 3.51\% & 60th - 80th & 2x \\
$>$ 3.51\% & Above 80th & 1x (No expansion) \\
\bottomrule
\end{tabular}
\end{table}

The use of external bounding boxes for the SFT stage is ablated in Appendix \ref{appendix:ll-crop-sft}, showing that \textbf{CropVLM} achieves similar performance even when initialized through a simpler procedure that does not involve an extra model.

\paragraph{Training Stages.} The Supervised Finetuning (SFT) stage trained 3 separate SmolVLM models at 3 different input resolutions, i.e. 512x512, 1024x1024 and 2048x2048 pixels, on the synthetic bounding box dataset. These SmolVLM models can produce rough bounding box extimates in response to the prompt: "\textit{\{QUESTION\} Outline the region in the image that would help answer this question.}". Then, the GRPO stage improved each of these models at the aforementioned resolutions on the remaining half of the dataset. In the GRPO stage, the generated bounding boxes are evaluated using a reward model that corresponds to a base 256M SmolVLM Instruct model with an input resolution of 512x512 pixels, motivated by computational efficiency. Using a lightweight model as the target VLM, from which rewards are derived, ensures fast evaluation while still being sufficiently accurate to provide meaningful learning signals, particularly because of group-normalized rewards. The experiments in Section \ref{sec:eval-cropvlm-w-other-vlms} show that \textbf{CropVLM}, framed with this reward model, can also be used to improve the results of larger VLMs. As described in Section \ref{sec:methods}, we experimented with 2 reward types, i.e. accuracy and likelihood-based, in addition to a bounding box format validation reward. The bounding box validation reward takes the value of 1 if the bounding box is valid and 0 otherwise, when training with likelihood-based rewards, and 0.25 if valid and 0 otherwise, for the accuracy-based rewards. This difference in scaling was necessary since the two types of rewards operate on different scales. The accuracy reward used each dataset's official metrics: VQA accuracy for TextVQA, and ANLS (Average Normalized Levenshtein Similarity) for DocVQA, ST-VQA, and InfographicVQA. The likelihood reward was calculated using the log-likelihood over the ground truth response tokens. In both cases, the prompt used in the reward model was the following: "\textit{\{QUESTION\} Give a very brief answer.}".

\paragraph{Hyperparameters.}
For all training runs, we employed LoRA \citep{hu2022lora} over the SmolVLM model with consistent parameters: rank 128, alpha 256, and dropout of 0.05. We maintained a batch size of 16 across all tests and used a cosine decay learning rate scheduler. All models were trained for a single epoch. Stage-specific hyperparameters varied as follows: For the SFT stage, we used a learning rate of 5e-5 and maximum gradient norm of 1.0. For the GRPO stage, we reduced these values to 5e-6 and 0.1, respectively. Additional GRPO-specific hyperparameters included a group size of 6, generation temperature of 0.8, and beta of 0.01. These conservative hyperparameters, namely the small group size in GRPO, together with the use of limited reinforcement learning data, and also small model sizes, were necessary due to resource constraints. Specifically, all our models were trained on a single A100 GPU. Using these hyperparameters, the SFT stage required approximately 3 GPU hours while the GRPO stage required 24 GPU hours for our largest resolution \textbf{CropVLM} model at 2048px. As a result, our reported results likely represent a lower bound on the full potential of the \textbf{CropVLM} method under training regimes that involve more resources.

\subsection{Experimental Results}

To assess the performance of our method, we evaluated on splits of the same datasets used during training. Specifically, for document and infographics understanding, we use the validation splits of DocVQA and InfographicsVQA. For scene-text analysis, we use the validation split of TextVQA and the test split of ST-VQA. To assess zero-shot high-resolution general vision understanding, we evaluate on the V* \cite{wu2024v} benchmark and also on the recent HR-Bench 4k and 8k \cite{wang2025divide} benchmarks. For each dataset, we employ the official evaluation metric: VQA accuracy for TextVQA, V* and HR-Bench 4k and 8k, and ANLS for DocVQA, ST-VQA and InfographicVQA. Appendices \ref{appendix:human-bbox}, \ref{appendix:ablations}, and \ref{sec:qual-examples} respectively contain statistics for bounding boxes generated by our method, ablations on the use of the full image, and qualitative examples obtained with \textbf{CropVLM}.

\subsubsection{Evaluating \textbf{CropVLM} Paired with SmolVLM}
\begin{table*}[htbp]
\centering
\begin{tabular}{lcccccccc}
\toprule

\textbf{Model} & \textbf{Resolution} & \textbf{Reward} & \textbf{TextVQA} & \textbf{DocVQA} & \textbf{InfoVQA} & \textbf{ST-VQA}  & \textbf{Average} \\
\midrule
SmolVLM & 512 & - & 39.49 & 13.68 & 13.08 & 47.53 & 28.45 \\
+ SFT & 512 & - & 43.55 & 20.23 & 14.86 & 50.39 & 32.26 \\
+ SFT + GRPO & 512 & Accuracy & \textbf{48.51} & 29.70 & \textbf{17.70} & 55.13 & 37.76 \\
+ SFT + GRPO & 512 & LL & 47.72 & \textbf{32.19} & 17.32 & \textbf{55.29} & \textbf{38.13} \\
\midrule
SmolVLM & 1024 & - & 52.71 & 47.86 & 20.12 & 57.49 & 44.55 \\
+ SFT & 1024 & - & 53.46 & 53.17 & 21.84 & 57.71 & 46.55 \\
+ SFT + GRPO & 1024 & Accuracy & 55.93 & 57.98 & 25.09 & 59.98 & 49.75 \\
+ SFT + GRPO & 1024 & LL & \textbf{56.94} & \textbf{58.75} & \textbf{26.59} & \textbf{61.26} & \textbf{50.89} \\
\midrule
SmolVLM & 2048 & - & 55.02 & 60.13 & 26.84 & 58.66 & 50.16 \\
+ SFT & 2048 & - & 52.29 & 60.89 & 26.62 & 57.37 & 49.29 \\
+ SFT + GRPO & 2048 & Accuracy & 55.82 & 61.85 & 29.76 & 60.56 & 52.00 \\
+ SFT + GRPO & 2048 & LL & \textbf{56.88} & \textbf{62.14} & \textbf{30.72} & \textbf{60.81} & \textbf{52.64} \\
\bottomrule
\end{tabular}

\caption{CropVLM performance when the model is paired with SmolVLM as the target model, across different resolutions. In this set of experiments, both the cropping network and the target model that responds to the requests operate at the same input resolution, as shown under \textbf{Resolution}. The label \textbf{Accuracy} refers to models trained with GRPO where the reward was the accuracy metric for each of the used datasets, while \textbf{LL} refers to models trained using the log-likelihood reward of the correct response. The \textbf{Average} column shows the average performance across datasets.}
\label{tab:cropvlm-results}
\end{table*}

We evaluate the impact of \textbf{CropVLM} on the SmolVLM 256M Instruct model across three input resolutions -- 512×512, 1024×1024, and 2048×2048 pixels -- and two training stages: Supervised Fine-Tuning (SFT) and Grouped Reward Policy Optimization (GRPO). In this evaluation, both the cropping network and the answering model process images at the same input resolution. Table~\ref{tab:cropvlm-results} summarizes the aforementioned VQA benchmarks.

\paragraph{Post-SFT Performance.}
After SFT, we observe resolution-dependent effects. Models at 512×512 and 1024×1024 resolution show consistent performance gains over their baselines, indicating that lower resolution \textbf{CropVLM} models learn useful patterns even during synthetic supervised training. In contrast, the 2048×2048 model sees a slight performance drop, particularly on the TextVQA and ST-VQA benchmarks, which suggests overfitting at higher resolutions during this stage.

\paragraph{Post-GRPO Performance.}
Applying GRPO results in substantial improvements across all resolutions and reward types. The 512×512 model has the largest relative gains, where cropping helps compensate for the model's lower perceptual capacity and visual fragmentation. Log-likelihood rewards generally outperform accuracy-based ones, likely due to the fact that they can provide more nuanced feedback during training. This option virtually eliminates examples where the reward is the same within a group, which in turn leads to more examples effectively contributing to model weight updates.


\paragraph{Key Insights.}
Several noteworthy patterns emerge from our evaluation results. Firstly, the 2048×2048 model underperforms after SFT but surpasses the baseline after GRPO, indicating that high-resolution models can still learn to leverage the cropping mechanism through reinforcement learning, even when trained using a low-resolution reward model (recall that \textbf{CropVLM} training always used SmolVLM with a resolution of 512x512 pixels for computing the reward). Additionally, the 1024×1024-pixel \textbf{CropVLM} model trained with log-likelihood rewards, when paired with a same-resolution SmolVLM model, achieves a higher average performance than the baseline 2048×2048 SmolVLM across benchmarks, despite relying on models operating at a lower resolution. Finally, improvements across all resolutions confirm that the cropping strategy provides valuable learning signals, independent of the target model's resolution.

\subsubsection{Evaluating CropVLM with Other VLMs}
\begin{table*}
    \centering
\begin{tabular}{lccccccccc}
\toprule
\textbf{Method} & \textbf{Resolution} & \textbf{TextVQA} &\textbf{DocVQA} & \textbf{InfoVQA} & \textbf{ST-VQA} & \textbf{V*} & \textbf{HR-4k} & \textbf{HR-8k} & \textbf{Average} \\
\midrule
LLaVA 1.5 & - & 48.03 & 23.28 & 20.70 & 52.48 & 42.41 & 35.25 & 34.65 & 36.69 \\
+ CropVLM & 512 & 52.25 & 26.40  & 21.95  & 55.21  & 46.07 & 39.88 & 35.88 & 39.66 \\
+ CropVLM & 1024 & 53.25  & 28.51  & 24.46   & 55.89   & 49.74 & \textbf{43.88} & 38.38 & 42.02 \\
+ CropVLM & 2048 & \textbf{53.93 } & \textbf{30.60 } & \textbf{25.61} & \textbf{56.81 } & \textbf{50.79} & 41.38 & \textbf{39.88} & \textbf{42.71} \\
\midrule
Qwen 2.5 VL & - & 68.20 & 69.52 & 44.03 & 65.49 & 51.31 & 51.88 & 44.50 & 56.42 \\
+ CropVLM & 512 & 75.46   & 77.88  & 49.80 & 68.57  & 56.02 & 64.75 & 63.00 & 65.07 \\
+ CropVLM & 1024 & \textbf{75.97} & 82.10  & 53.45 & \textbf{69.09} & \textbf{60.73} & 64.75 & \textbf{63.63} & 67.10 \\
+ CropVLM & 2048 & 75.72 & \textbf{84.41} & \textbf{55.95} & 68.31 & 59.69 & \textbf{65.13} & 60.75 & \textbf{67.14} \\
\midrule
GPT 4.1 nano & - & 53.38 & 48.16 & 25.43 & 53.28 & 32.98 & 38.75 & 36.88 & 41.27 \\
+ CropVLM & 512 & 54.79   & 52.32  & 29.81 & 53.96  & 38.22 & 38.63 & 38.88 & 43.80 \\
+ CropVLM & 1024 & \textbf{56.25} & 56.40  & 32.79    & 55.02   & 37.70 & 41.38 & 39.50 &  45.58 \\
+ CropVLM & 2048 & 55.68 & \textbf{58.17} & \textbf{36.85} & \textbf{55.15} & \textbf{42.41} & \textbf{43.13} & \textbf{40.50} & \textbf{47.41} \\
\bottomrule
\end{tabular}
    \caption{Performance of LLaVA 1.5 7B, Qwen 2.5 VL 3B, and  GPT 4.1 nano, paired with CropVLM across different input resolutions for the cropping network, with "-" in the column labeled as \textbf{Resolution} representing the standalone baseline model, without CropVLM. LLaVA 1.5 uses an input resolution of 336x336 pixels, Qwen 2.5 VL uses a maximum input resolution of 448x448 pixels, and  GPT 4.1 nano used an input resolution of 512x512 pixels. All CropVLM models were trained using the log-likelihood reward. The \textbf{Average} column shows the average performance across datasets.}
    \label{tab:cropvlm-other-baselines}
\end{table*}

\label{sec:eval-cropvlm-w-other-vlms}
\textbf{CropVLM} can also be paired with other VLMs. Specifically, we evaluated \textbf{CropVLM} at multiple input resolutions, trained with a log-likelihood reward as this was shown to be the best performing reward type, with the following target VLMs: LLaVA 1.5 7B at 336x336 pixels input resolution; Qwen 2.5 VL 3B at 448x448 maximum input pixels resolution; and GPT 4.1 nano at 512x512 pixels of input resolution and with the text generation temperature set at zero. Table \ref{tab:cropvlm-other-baselines} presents the obtained results.

As expected, cropping networks trained at higher resolutions tend to outperform their lower-resolution counterparts across models and datasets. There is also a larger relative improvement in the in-domain datasets that have smaller relative areas of interest in their images, e.g. DocVQA and InfographicVQA. We also observe improvements in the V* benchmark and the HR-Bench 4k and 8k benchmarks, providing evidence of strong out-of-distribution performance, both in terms of the cropping network and the answering model. Particularly, when prompting the baseline GPT 4.1 nano model, we often saw a refusal to answer the questions in V*, with the model stating that the objects in the query did not appear in the image. This decreased as the model was paired with higher resolution \textbf{CropVLM}s.\footnote{The baseline GPT 4.1 nano model refused to answer 31 out of the 191 questions in the V* benchmark, dropping to 7 and 2 questions when paired with \textbf{CropVLM} at 512 and 2048 pixels resolution, respectively.} \textbf{CropVLM} allows for optimizing external models without fine-tuning them or even having access to their weights. Appendix \ref{appendix:benefits-higher-res} shows VQA performance improvements at much larger resolutions.

\begin{table*}
    \centering
\begin{tabular}{lcccccccc}
\toprule
\textbf{Method} & \textbf{TextVQA} &\textbf{DocVQA} & \textbf{InfoVQA} & \textbf{ST-VQA} & \textbf{V*} & \textbf{HR-4k} & \textbf{HR-8k} & \textbf{Average} \\
\midrule
LLaVA 1.5 & 48.03 & 23.28 & 20.70 & 52.48 & 42.41 & 35.25 & 34.65 & 36.69 \\
+ ViCrop (\texttt{rel-att}) & 55.22 & 25.96 & 21.23 & 56.95 & 49.21 & \textbf{44.25} & 36.50 & 41.33 \\
+ ViCrop (\texttt{grad-att}) & \textbf{56.19} & 25.84 & 21.43 & 57.06 & 48.17 & 43.62 & 37.12 & 41.35 \\
+ UV-CoT & 53.46 &  28.47 & 21.07 & \textbf{59.30} & 43.98 & 37.88 & 35.88 & 40.01 \\
+ CropVLM & 53.93 & \textbf{30.60} & \textbf{25.61} & 56.81 & \textbf{50.79} & 41.38 & \textbf{39.88} & \textbf{42.71} \\
\midrule
Qwen 2.5 VL & 68.20 & 69.52 & 44.03 & 65.49 & 51.31 & 51.88 & 44.50 & 56.42 \\
+ ViCrop (\texttt{rel-att}) & 72.91 & 71.40 & 47.94 & \textbf{68.96} & 53.40 & 56.25 & 47.38 & 59.75 \\
+ ViCrop (\texttt{grad-att}) & 74.15 & 72.27 & 49.43 & 68.09 & 53.40 & 54.38 & 46.00 & 59.67 \\
+ UV-CoT & 74.56 & 76.60 & 47.98 & 67.91 & 56.54 & 53.62 & 47.25 & 60.64 \\
+ CropVLM & \textbf{75.72} & \textbf{84.41} & \textbf{55.95} & 68.31 & \textbf{59.69} & \textbf{65.13} & \textbf{60.75} & \textbf{67.14} \\
\bottomrule
\end{tabular}
    \caption{Comparison of our CropVLM approach against other similar approaches in the literature. The CropVLM model present in this table receives as input an image at 2048x2048 pixels and was trained using a log-likelihood reward. Both LLaVA 1.5 and UV-CoT operate at 336x336 pixels of input resolution, and Qwen 2.5 VL operates at 448x448 pixels of input resolution. The \textbf{Average} column shows the average performance across datasets.}
    \label{tab:cropvlm-vs-vicrop}
\end{table*}

\subsubsection{Comparing the \textbf{CropVLM} Model Against Alternative Cropping Strategies}

We compared our \textbf{CropVLM} strategy against two alternative methods: ViCrop \cite{zhang2025mllms} and UV-CoT \cite{zhao2025unsupervised}. ViCrop is a training-free approach that uses explainability techniques, based on gradients and attention weights, to identify regions of interest within an image, given a query. We evaluate ViCrop using the two best-performing techniques, as reported in the original paper - relative attention (\texttt{rel-attn}) and gradient-weighted attention (\texttt{grad-attn}) - implementing them with the officially released code and recommended hyperparameters, with two base VLMs respectively employing LLaVA 1.5 7B at 336x336 pixels, and Qwen 2.5 at 448x448 pixels resolution, to both generate bounding boxes and answer questions. UV-CoT, on the other hand, is a DPO based training method to learn both a relevant crop and the correct response to a request given an image. We use the released model\footnote{\url{https://huggingface.co/kesenZhaoNTU/UV-CoT}} to produce relevant crops, and pair it with the same answering models as ViCrop. In turn, \textbf{CropVLM} processes inputs at 2048×2048 pixels resolution, while the answering models receive images at the same resolution as their alternative counterparts. Note that VisRL \cite{chen2025visrl} was not available at the time of writing and therefore was not included in our evaluation. Results are presented in Table \ref{tab:cropvlm-vs-vicrop}.

\textbf{CropVLM} outperforms ViCrop across most datasets, with particularly notable advantages on DocVQA and InfographicVQA when paired with LLaVA 1.5. This performance gap can be attributed to the fact that these datasets can be considered as out-of-domain for LLaVA 1.5, which reduces the reliability of ViCrop's explainability methods. Conversely, ViCrop performs relatively better on datasets that are closer to the underlying model’s training distribution, such as TextVQA and ST-VQA when applied on LLaVA 1.5. Our method also demonstrates superior performance on V* and HR-Bench 8k with LLaVA 1.5, which is significant because these datasets are out-of-domain for both \textbf{CropVLM} and the target model. When paired with Qwen 2.5 VL, \textbf{CropVLM} significantly outperforms ViCrop on most datasets, with ST-VQA being the only exception. However, as shown in Table \ref{tab:cropvlm-other-baselines}, the 1024×1024 pixel resolution version of \textbf{CropVLM} does surpass ViCrop on ST-VQA when using Qwen 2.5 VL. Regarding UV-CoT, our method outperforms it on all datasets across both answering models, with the sole exception of ST-VQA when paired with LLaVA 1.5. This highlights the efficiency of our approach, as \textbf{CropVLM} achieves stronger overall performance despite being smaller - 256M parameters compared to UV-CoT’s 7B - and using less training data - 62k VQA examples in our case and 249k preference pairs for UV-CoT. Additionally, Appendix \ref{appendix:our-viscot} presents an analysis of a cropping network trained via supervised fine-tuning on Visual-CoT data, demonstrating that models of our scale struggle to learn effective cropping strategies for VQA tasks.

\subsection{Execution Overhead}

\begin{table}
\small
\centering
\begin{tabular}{lrrr}
\toprule
\textbf{Method} & \textbf{Res.} & \textbf{Time} & \textbf{Memory} \\
\midrule
LLaVA 1.5 (\texttt{rel-att})  & 336 & 301.0 & 21158 \\
LLaVA 1.5 (\texttt{grad-att})  & 336 & 260.8 & 21199 \\
Qwen 2.5 VL (\texttt{rel-att})  & 448 & 396.1 & 19281 \\
Qwen 2.5 VL (\texttt{grad-att}) & 448 & 525.3 & 19294 \\
\midrule
UV-CoT & 336 & 727.2 & 15252 \\
\midrule
CropVLM & 2048 & 691.1 & 1738 \\
CropVLM & 1024 & 495.7 & 1256 \\
CropVLM & 512 & 418.3 & 1164 \\
\bottomrule
\end{tabular}
\caption{Execution overhead of CropVLM and alternative cropping methods, measured on the same 100 random samples from the TextVQA dataset. \textbf{Time} denotes the duration, in milliseconds, required to generate a bounding box, while \textbf{Memory} refers to the peak GPU memory usage in megabytes. \textbf{Res.} denotes the input resolution uses in each model.}
\label{tab:execution-overhead}
\end{table}

In Table~\ref{tab:execution-overhead}, we report the average execution time overhead, and the peak GPU memory usage of \textbf{CropVLM}, comparing these results to those of ViCrop and UV-CoT, as measured on the same 100 random samples from the TextVQA dataset. All measurements were taken on a machine equipped with an NVIDIA RTX A6000 GPU and an Intel(R) Xeon(R) Gold 6348 CPU @ 2.60GHz. \textbf{CropVLM} demonstrates significantly lower GPU memory usage, largely due to the smaller model size and its compatibility with FlashAttention \citep{dao2023flashattention}. In contrast, ViCrop requires access to the full attention scores, making it incompatible with FlashAttention and resulting in higher memory consumption. On the other hand, \textbf{CropVLM} is slower in generating bounding boxes because it relies on text generation to produce coordinate values, while ViCrop does not. The underlying model (i.e., SmolVLM 256M Instruct) has a highly restricted numeric output vocabulary, limited to the digits 0 through 9. As a result, generating numbers between 0 and 100 requires emitting multiple tokens per number, which increases decoding time. UV-CoT, while also compatible with FlashAttention, is substantially larger and also requires generating multiple tokens for a single bounding box, which makes it slower and more memory-demanding than \textbf{CropVLM}, even at lower resolutions. Future work can perhaps explore alternative approaches to the design of the cropping network.

\section{Conclusions}
\label{sec:conclusion}

We introduce \textbf{CropVLM}, a lightweight cropping network that improves VLM performance on fine-grained image understanding tasks. \textbf{CropVLM} operates externally and requires no access to model weights, making it compatible with both open-source and proprietary VLMs. It is trained using reinforcement learning with objective reward signals - such as the log-likelihood of the correct answer - that directly measure how informative a crop is, grounded directly in the language modeling behavior of modern VLMs. This approach differs from prior methods that rely on general-purpose models for reward attribution, providing a more task-aligned and quantifiable assessment of crop utility. It outperforms the crops generated by UV-CoT, as well as methods like ViCrop, especially in out-of-domain settings where explainability-based signals break down. \textbf{CropVLM} delivers consistent gains with low computational overhead, making it a scalable and general solution for enhancing VLMs on high-resolution, detail-sensitive tasks.



\section*{Acknowledgements}
This research was supported by the Portuguese Recovery and Resilience Plan through project C645008882-00000055 (i.e., the Center For Responsible AI), and also by the Fundação para a Ciência e a Tecnologia, I.P. (FCT) under the projects with references UID/50021/2025 (DOI: https://doi.org/10.54499/UID/50021/2025) and UID/PRR/50021/2025 (DOI: https://doi.org/10.54499/UID/PRR/50021/2025).

{
    \small
    \bibliographystyle{ieeenat_fullname}
    \bibliography{main}
}

\appendix
\section{Limitations and Ethical Considerations}
\label{appendix:limitations}

The research reported in this paper aims to refine the capabilities of VLMs by enabling detailed visual understanding without the computational costs associated with uniformly increasing input resolution. This objective aligns with broader goals related to green AI and transparent reporting of the limitations of current VLMs. However, there remain several important considerations for responsible development and deployment.

A first limitation of our study concerns the scope of the experimental setting. Our experiments rely exclusively on English-language models and datasets, which restricts our ability to assess multilingual generalization. Moreover, extending this line of work to multilingual VLMs would require carefully designed mixtures of multilingual VQA data for both training and evaluation.

Another key consideration involves the well-known biases present in VLMs, inherited from large-scale pre-training corpora or from the use of foundation encoders such as CLIP. While we conduct our experiments on publicly available VQA datasets commonly used in prior work, we do not analyze how automatically selected image regions may amplify or suppress particular biases. Caution is therefore warranted before deploying systems such as ours in sensitive real-world environments, and we highlight the need for future research on fairness-oriented evaluation in cropping-based VLM pipelines.

Overall, our approach should be viewed as a step toward more efficient fine-grained visual reasoning, rather than a complete solution to the broader challenges surrounding robustness, fairness, and multilingual accessibility in VLMs.

\section{Evaluating the Generated Crops Against Human-annotated Bounding Boxes}
\label{appendix:human-bbox}

\begin{table*}
    \centering
    \begin{tabular}{lccccccc}
    \toprule
    \textbf{Model} & \textbf{Resolution}  &\textbf{IoU} & \textbf{Recall} & \textbf{Full Recall} & \textbf{Size} & \textbf{TextVQA} \\
    \midrule
    CropVLM SFT & 512 & 14.85 & 44.16 & 19.41 & 14.36 & 43.26 \\
    CropVLM Accuracy & 512 &15.52 & 76.34 & 52.01 & 26.94 & 48.41 \\
    CropVLM LL & 512 & 10.91 & 87.95 & 69.95 & 45.48 & 47.55 \\
    \midrule
    CropVLM SFT & 1024 & 17.85 & 51.37 & 22.56 & 12.47 & 53.49 \\
    CropVLM Accuracy & 1024 & 18.61 & 71.95 & 44.92 & 16.81 & 56.05 \\
    CropVLM LL & 1024 & 13.46 & 84.99 & 64.30 & 33.39 & 57.14 \\
    \midrule
    CropVLM SFT & 2048 & 17.90 & 52.41 & 22.84 & 13.15 & 52.16 \\
    CropVLM Accuracy & 2048 & 18.07 & 74.58 & 47.43 & 17.83 & 55.87 \\ 
    CropVLM LL & 2048 & 14.51 & 84.41 & 63.56 & 29.22 & 56.96 \\
    \midrule
    \midrule
    ViCrop LLaVA 1.5 (\texttt{rel-attn}) & 336 & 13.70 & 71.27 & 46.36 & 17.20 & 55.11 \\
    ViCrop LLaVA 1.5 (\texttt{grad-attn}) & 336 & 13.58 & 73.45 & 48.40 & 18.07 & 56.07 
    \\
    \midrule
    ViCrop Qwen2.5 VL (\texttt{rel-attn}) & 448 & 16.36 & 55.21 & 29.08 & 7.48 & 72.92 \\
    ViCrop Qwen2.5 VL (\texttt{grad-attn}) & 448 & 17.74 & 62.07 & 33.62 & 7.44 & 74.41 \\
    \midrule
    \midrule
    UV-CoT & 336 & 14.73 & 52.61 & 18.58 & 10.75 & 53.33 \\
    \bottomrule
    
    \end{tabular}

\caption{Bounding box quality metrics and corresponding TextVQA performance on a human-annotated subset. CropVLM models are denoted by training stage, where \textbf{SFT} represents supervised fine-tuning before reinforcement learning, \textbf{Accuracy} represents GRPO performed with accuracy rewards, and \textbf{LL} represents  GRPO performed with log-likelihood rewards. \textbf{Resolution} denotes the input resolution used in each model. ViCrop models are denoted by the model being used and the method. UV-CoT is used here as a cropping network only. TextVQA performance reflects both cropping quality and the underlying answering model (SmolVLM for CropVLM, LLaVA 1.5/Qwen2.5 VL for ViCrop, LLaVA 1.5 for UV-CoT), making direct performance comparisons not meaningful across different answering models.}
\label{tab:bbox-iou}
\end{table*}

The ViCrop authors released human-given bounding box annotations for a subset of the TextVQA dataset, consisting of 4370 samples where each image contains a single annotated region of interest. We made use of this dataset to analyze the bounding boxes produced with \textbf{CropVLM}, considering it instead of the VisCoT benchmark since the latter relies on synthetic annotations generated via PaddleOCR for text-centric datasets. Using this data, we calculate the following statistics to describe the behavior of our method, in comparison with ViCrop and UV-CoT. Below, $B_p$ is the predicted bounding box and $B_{gt}$ represents the ground-truth bounding box for each instance.

\begin{itemize}
    \item \textbf{Intersection over Union (IoU):} Measures the relative area of overlap between $B_p$ and $B_{gt}$.
    \item \textbf{Recall:} Percentage of the $B_{gt}$ area that is covered by $B_p$.
    \item \textbf{Full Recall:} Percentage of instances where $B_p$ fully contains $B_{gt}$.
    \item \textbf{Average Relative Bounding Box Size (Size):} Ratio between $B_p$ and the total image area.
\end{itemize}

Results are presented in Table~\ref{tab:bbox-iou}. Notably, bounding boxes generated after GRPO training are significantly larger than those from the SFT stage, regardless of reward type, for \textbf{CropVLM} models. This increase in bounding box size, following the reinforcement learning step, suggests that the models learn to capture broader regions to ensure comprehensive information coverage.

A counter-intuitive finding emerges when examining IoU metrics in relation to task performance. IoU does not correlate with TextVQA accuracy when the cropping networks use the same target VLM. For example, among \textbf{CropVLM} models at 1024 pixels resolution, the SFT model achieves an IoU of 17.85 but lowest TextVQA performance (53.49), while the LL model has the lowest IoU (13.46) but highest performance (57.14). Similarly, when comparing UV-CoT with ViCrop LLaVA 1.5 methods, higher IoU does not translate to better performance. This suggests that tighter alignment with human annotations does not necessarily correlate with improved question-answering performance. In contrast, recall and full recall metrics, when considered together with relative bounding box size, provide better signals for predicting TextVQA performance within the same answering model.

Additionally, \textbf{CropVLM} models trained with reinforcement learning tend to generate progressively smaller bounding boxes as input resolution increases, likely because higher-resolution inputs provide sufficient detail for the model to confidently identify and "zoom in" on specific regions rather than capturing broader areas. This progressive refinement aligns with the expected behavior of an effective cropping mechanism.

\section{Benefits at Higher Resolutions}
\label{appendix:benefits-higher-res}
We tested our \textbf{CropVLM} model at a 2048$\times$2048 input resolution, pairing it with Qwen 2.5 VL 3B, whose maximum supported resolution is 1792$\times$1792. This setup represents roughly a 16$\times$ increase in pixel count compared to the settings used for the results in Tables 4 and 5 of the main paper. The results, shown in Table~\ref{tab:cropvlm-high-res}, indicate that \textbf{CropVLM continues to provide improvements across most benchmarks}, particularly on datasets that are in-domain with respect to the cropping network’s training data. This suggests that visual fragmentation remains an issue even at higher input resolutions, and that our method still helps reduce it. However, the gains are less consistent on V* and HR-Bench 4k/8k, suggesting that for models operating at very high resolutions, the benefits of external cropping may diminish in settings that are out-of-distribution for the cropping model.

\begin{table*}[htbp]
    \centering
\begin{tabular}{lccccccccc}
\toprule
\textbf{Method} & \textbf{Reward} &  \textbf{TextVQA} &\textbf{DocVQA} & \textbf{InfoVQA} & \textbf{ST-VQA} & \textbf{V*} & \textbf{HR-4k} & \textbf{HR-8k} & \textbf{Average}\\
\midrule
Qwen 2.5 VL & - & 79.14 & 92.50 & 75.55 & 66.67 & 73.30 & \textbf{67.75} &  63.88 & 74.11 \\
+ CropVLM & Accuracy & \textbf{80.12} & \textbf{92.83} & \textbf{75.60} & 68.71 &  72.25 & 65.75 & \textbf{65.63} & 74.41 \\
+ CropVLM & LL & 80.07 & \textbf{92.83} & \textbf{75.60} & \textbf{68.72} &  \textbf{74.35} & 66.38 & 63.25 & \textbf{74.46}\\
\bottomrule
\end{tabular}
    \caption{
   Performance of a Qwen 2.5 VL model, operating at 1792x1792 pixels input resolution, paired with \textbf{CropVLM}. The \textbf{CropVLM}
model processes images at 2048x2048 pixels of resolution. The \textbf{Average} column shows the
average performance across datasets.}
    \label{tab:cropvlm-high-res}
\end{table*}

\section{Ablating the Use of the Full Image}
\label{appendix:ablations}


To evaluate the importance of accessing the complete image context when responding to requests, we conducted an ablation study comparing  \textbf{CropVLM}'s performance when paired with SmolVLM, with and without the full image also being made available, using models trained with accuracy-based rewards and log-likelihood rewards. The results are presented in Table \ref{tab:ablations}. We observe that access to the complete image significantly enhances the model's ability to generate accurate responses, as there is a performance decrease for the version with only crops, across all datasets. Notably, in InfographicVQA, the performance of the model with just a crop still exceeds the baseline model without  \textbf{CropVLM} for both reward types.

\begin{table*}[tbp]
\centering
\begin{tabular}{lcccccc}
\toprule
\textbf{Model} & \textbf{Reward} & \textbf{TextVQA} & \textbf{DocVQA} & \textbf{InfoVQA} & \textbf{ST-VQA} & \textbf{Average} \\
\midrule
SmolVLM & - & 55.02 & 60.13 & 26.84 & 58.66 & 50.16 \\
+ CropVLM (\textbf{w/o full image}) & Accuracy & 49.13 & 53.27 & 27.03 & 53.93 & 45.84 \\
+ CropVLM & Accuracy & 55.82 & 61.85 & 29.76 & 60.56 & 52.00 \\
+ CropVLM (\textbf{w/o full image}) & LL & 53.37 & 54.72 & 29.70 & 56.62 & 48.60 \\
+ CropVLM & LL & \textbf{56.88} & \textbf{62.14} & \textbf{30.72} & \textbf{60.81} & \textbf{52.64} \\
\bottomrule
\end{tabular}
\caption{Performance of SmolVLM paired with CropVLM, with both models operating at 2048x2048 pixels of input resolution, with and without the full image. The \textbf{Average} column reports the average performance across datasets.}
\label{tab:ablations}
\end{table*}

\section{Learning to Crop without External Help}
\label{appendix:ll-crop-sft}

Our \textbf{CropVLM} models were trained in two stages: initial Supervised Fine-Tuning (SFT) to teach the model to generate valid bounding boxes, followed by GRPO to refine its outputs. To train a model like SmolVLM to generate valid bounding boxes, we first applied SFT using data generated by Qwen 2.5 VL, with slight modifications. Specifically, we expanded smaller bounding boxes to increase the likelihood that they contained the relevant image regions.  

To evaluate the impact of external data on \textbf{CropVLM}, we repeat the experiments using bounding boxes generated through an alternative method base on exhaustive search. In this setup, we create all possible crops within an $N \times N$ grid for each image, with $N$ equal to 5, and evaluate the log-likelihood of the correct response given the full image, the candidate crops, and the question. SmolVLM, operating at a resolution of $512 \times 512$ pixels for efficiency, selects the bounding box corresponding to the highest log-likelihood as the target for SFT.  

Additionally, to ensure the model can produce bounding boxes covering the full coordinate range (0–100), we add random perturbations to the box coordinates. For each training example, noise is sampled uniformly from $[0, \frac{100}{N}]$ and subtracted from the upper-left corner while being added to the lower-right corner, expanding the box outward from its center while keeping all coordinates valid. The reinforcement learning stage is performed as before.

Results are presented in Table \ref{tab:ll-crop} demonstrating that \textbf{CropVLM} models trained without external bounding box supervision achieve performance similar to those trained with Qwen 2.5 VL supervision. We also observe a larger performance gain from the SFT-only stage to the GRPO stage when supervision does not rely on external data. This underscores both the robustness of our approach and the critical role of reinforcement learning in producing relevant bounding box predictions.

\begin{table*}[htbp]
\centering

\begin{tabular}{lccccccc}
\toprule
\textbf{Model} & \textbf{Resolution} & \textbf{External} & \textbf{TextVQA} & \textbf{DocVQA} & \textbf{InfoVQA} & \textbf{ST-VQA} & \textbf{Average} \\
\midrule
SmolVLM & 512 & - & 39.49 & 13.68 & 13.08 & 47.53 & 28.45 \\
+ SFT & 512 & \checkmark & 43.55 & 20.23 & 14.86 & 50.39 & 32.26 \\
+ SFT & 512 & \xmark & 39.05 & 21.06 & 13.65 & 47.05 & 30.20 \\
+ SFT + GRPO & 512 & \checkmark & \textbf{47.72} & 32.19 & \textbf{17.32} & \textbf{55.29} & \textbf{38.13} \\
+ SFT + GRPO & 512 & \xmark & 45.83 & \textbf{33.47} & 16.76 & 53.90 & 37.49 \\
\midrule
SmolVLM & 1024 & - & 52.71 & 47.86 & 20.12 & 57.49 & 44.54 \\
+ SFT & 1024 & \checkmark & 53.46 & 53.17 & 21.84 & 57.71 & 46.54 \\
+ SFT & 1024 & \xmark & 52.32 & 52.20 & 23.42 & 57.52 & 46.36 \\
+ SFT + GRPO & 1024 & \checkmark & \textbf{56.94} & \textbf{58.75} & \textbf{26.59} & 61.26 & \textbf{50.89} \\
+ SFT + GRPO & 1024 & \xmark & 55.97 & 57.34 & 26.43 & \textbf{61.34} & 50.27 \\
\midrule
SmolVLM & 2048 & - & 55.02 & 60.13 & 26.84 & 58.66 & 50.16 \\
+ SFT & 2048 & \checkmark & 52.29 & 60.89 & 26.62 & 57.37 & 49.29 \\
+ SFT & 2048 & \xmark & 53.66 & 57.71 & 28.66 & 58.48 & 49.63 \\
+ SFT + GRPO & 2048 & \checkmark & \textbf{56.88} & \textbf{62.14} & 30.72 & 60.81 & \textbf{52.64} \\
+ SFT + GRPO & 2048 & \xmark & 56.46 & 59.86 & \textbf{31.36} & \textbf{61.24} & 52.23 \\
\bottomrule
\end{tabular}
\caption{Performance of \textbf{CropVLM} paired with SmolVLM across different input resolutions. Both the cropping network and the target model operate at the same resolution. Models marked with \checkmark~in the \textbf{External} column were trained using Qwen 2.5 VL bounding boxes during SFT, while models marked with \xmark~were trained using noised crops obtained through exhaustive log-likelihood search. All models undergoing GRPO were trained with log-likelihood rewards. The \textbf{Average} column reports the average performance across datasets.}
\label{tab:ll-crop}
\end{table*}

\section{Training a Cropping Network with Tight Bounding Boxes}
\label{appendix:our-viscot}

Using a subset of the Visual-CoT training data, which contains tight bounding box annotations for regions of interest in image–question pairs, we train a cropping network to predict regions of interest. This training is performed using supervised fine-tuning with the same hyperparameters as reported in Section 4, without modifying the original bounding box annotations. Specifically, we use the subsets of Visual-CoT corresponding to TextVQA, TextCaps, DocVQA, and InfographicsVQA, totaling 99k samples. In contrast, our GRPO stage used only 62k samples. Table \ref{tab:our-viscot} reports VQA performance, while Table \ref{tab:our-viscot-bbox-iou} reports detection performance on the TextVQA subset described in \ref{appendix:human-bbox}.

\begin{table*}[htbp]
\centering

\begin{tabular}{lcccccc}
\toprule
\textbf{Model} & \textbf{Reward} & \textbf{TextVQA} & \textbf{DocVQA} & \textbf{InfoVQA} & \textbf{ST-VQA} & \textbf{Average} \\
\midrule
SmolVLM  & - & 55.02 & 60.13 & 26.84 & 58.66 & 50.16 \\
+ CropVLM  & Accuracy & 55.82 & 61.85 & 29.76 & 60.56 & 52.00 \\
+ CropVLM  & LL & \textbf{56.88} & \textbf{62.14} & \textbf{30.72} & \textbf{60.81} & \textbf{52.64} \\
+ Visual-CoT  &  - & 50.53 & 59.25 & 25.72 & 56.23 & 47.93 \\
\bottomrule
\end{tabular}
\caption{SmolVLM paired with \textbf{CropVLM} and the model trained on a subset of the Visual-CoT training data, as denoted by \textbf{Visual-CoT}. Here, both the cropping network and the model that responds to the question operate at 2048 pixels resolution. The label \textbf{Accuracy} refers to models trained with GRPO where the reward was the accuracy metric for each of the used datasets, while \textbf{LL} refers to models trained using the log-likelihood reward of the correct response. The \textbf{Average} column shows the average performance across datasets.}
\label{tab:our-viscot}
\end{table*}

\begin{table*}[htbp]
    \centering
    \begin{tabular}{lccccc}
    \toprule
    \textbf{Model} & \textbf{IoU} & \textbf{Recall} & \textbf{Full Recall} & \textbf{Size} & \textbf{TextVQA} \\
    \toprule
    CropVLM - SFT & 17.90 & 52.41 & 22.84 & 13.15 & 52.16 \\
    CropVLM - Accuracy & 18.07 & 74.58 & 47.43 & 17.83 & 55.87 \\ 
    CropVLM - LL & 14.51 & 84.41 & 63.56 & 29.22 & 56.96 \\
    Visual-CoT & 14.67 & 17.23 & 0.23 & 1.27 & 50.24 \\
    \bottomrule
    \end{tabular}

\caption{Bounding box quality metrics and corresponding TextVQA performance on a human-annotated subset of the dataset. \textbf{CropVLM} cropping networks are denoted by training stage, where \textbf{SFT} represents supervised fine-tuning before reinforcement learning, \textbf{Accuracy} represents GRPO performed with accuracy rewards, and \textbf{LL} represents  GRPO performed with log-likelihood rewards. All cropping networks are paired with SmolVLM, and both models operate at 2048x2048 pixels input resolution. \textbf{Visual-CoT} denotes a model trained on a subset of the Visual-CoT training data.}
\label{tab:our-viscot-bbox-iou}
\end{table*}

Results show that a cropping network of this size cannot effectively learn to produce useful bounding boxes for VQA, that tightly enclose the region of interest for each image–question pair. While the model can generate bounding boxes of a similar size, it consistently fails to position them in the correct locations during testing, as reflected in the recall and VQA accuracy metrics. Moreover, even though the \textbf{CropVLM - LL} model achieves an IoU similar to that of the Visual-CoT–trained cropping network, its VQA performance is significantly lower: when paired with SmolVLM, it underperforms all our \textbf{CropVLM} variants and even the baseline. This further demonstrates that IoU does not correlate strongly with VQA accuracy and should not be used as a proxy metric for VQA performance. Interestingly, our reinforcement-learned \textbf{CropVLM} tends to predict substantially larger boxes, suggesting that when precise localization is difficult, the model compensates by expanding the region it selects to achieve higher recall.
\section{Expanding the Size of the Human-annotated Bounding Boxes}
\label{appendix:expanding-tight}

Datasets with human-annotated bounding boxes, such as the one introduced by the authors of ViCrop, typically provide tight crops around the regions of interest. While precise, these bounding boxes may exclude surrounding visual context that can be crucial for answering certain questions in VQA tasks. To investigate this, we evaluate VQA performance on the TextVQA human-annotated subset under different bounding box expansion factors. As in our main evaluation protocol for other models, we provide both the full image and the corresponding human-annotated crop to the target model. The bounding box centers are kept fixed, while their width and height are adjusted in this test, so that the area becomes a scaled multiple of the original, according to an expansion factor.

\begin{figure}[htbp]
    \centering
    \includegraphics[width=\columnwidth]{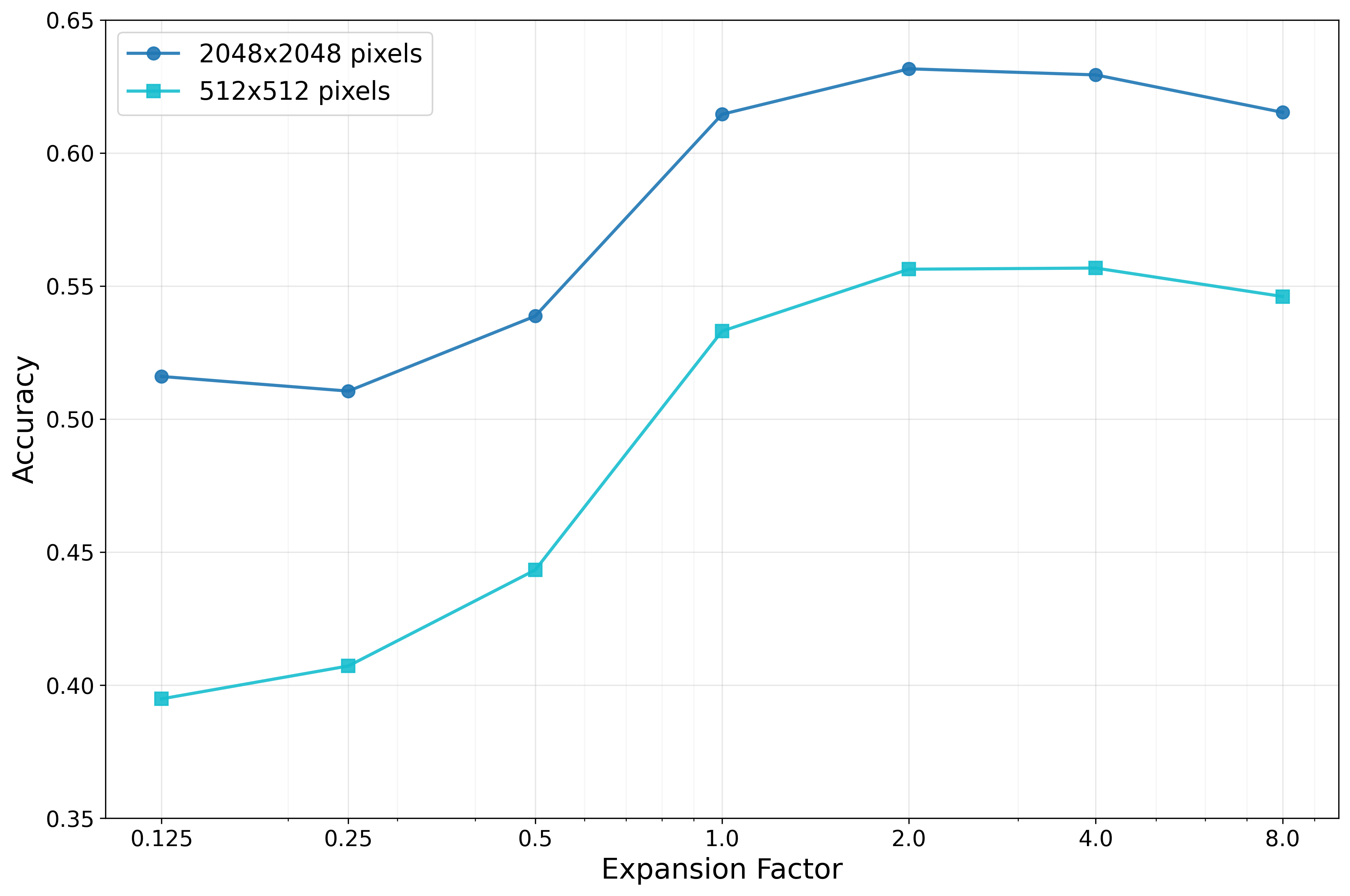}
    \caption{TextVQA performance across multiple bounding box expansion factors, using human-annotated annotations, and with SmolVLM at 512$\times$512 and 2048$\times$2048 input resolutions.}
    \label{fig:expanding-bbox}
\end{figure}

As shown in Figure~\ref{fig:expanding-bbox}, shrinking the bounding boxes immediately reduces performance, confirming that the original annotations are tightly focused on relevant regions. Interestingly, modest expansions can improve performance, suggesting that the tight boxes do not represent an upper bound and that additional surrounding context can be beneficial.

\section{CropVLM Qualitative Examples}
\label{sec:qual-examples}

\begin{figure*}
    \centering
    \includegraphics[width=1\textwidth]{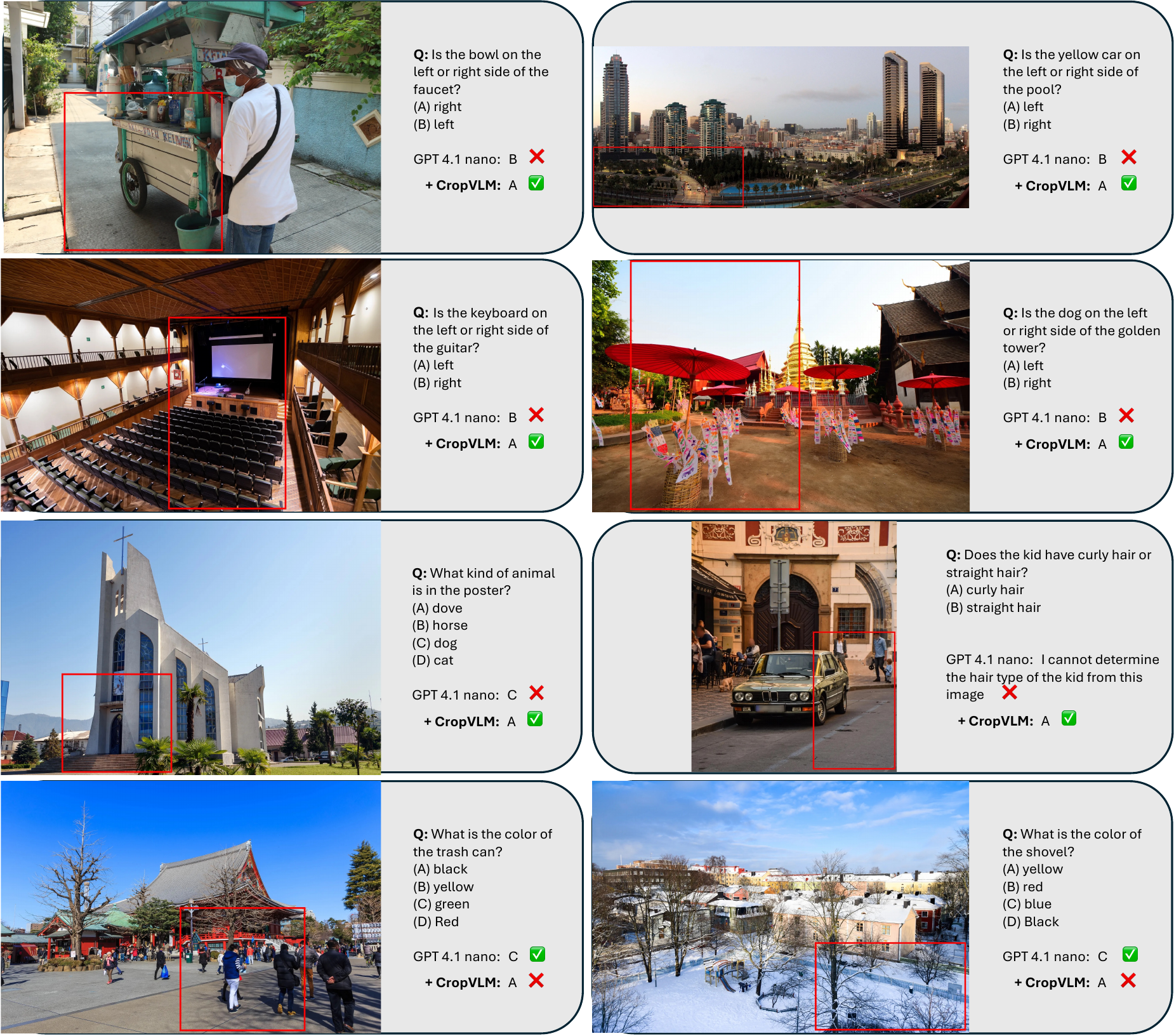}
    \caption{Qualitative examples from the V* Benchmark, where the first 6 cases are successful and the last 2 are failures. Next to each image, we present the question alongside responses from GPT 4.1 nano, and GPT 4.1 nano paired with the \textbf{CropVLM} model that accepts images at 2048x2048 pixels of input resolution, and which was trained using log-likelihood rewards. The red bounding box denotes the \textbf{CropVLM} proposed region of interest.}
    \label{fig:v-star-qual-examples}
\end{figure*}

\begin{figure*}
    \centering
    \includegraphics[width=1\textwidth]{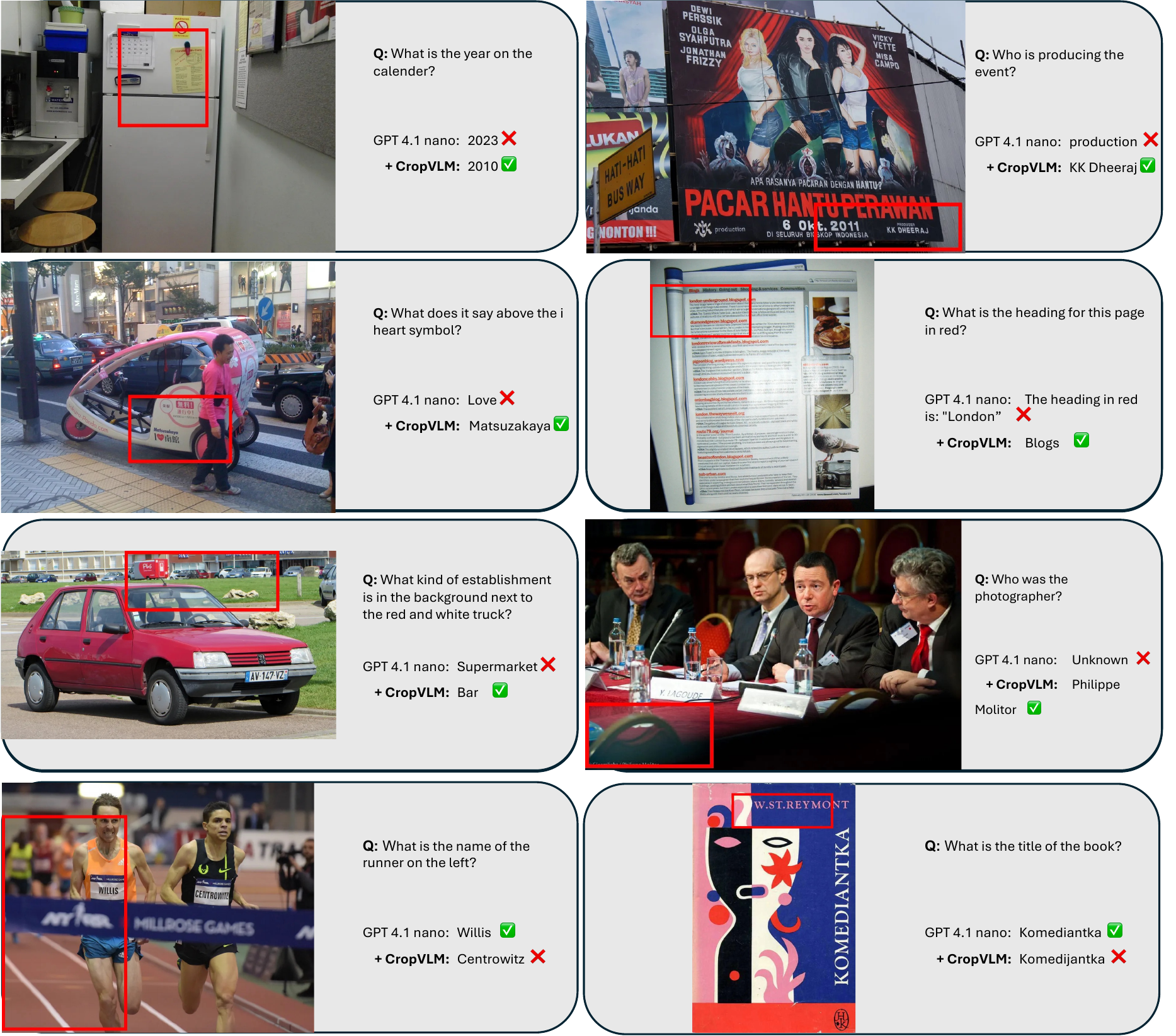}
    \caption{Qualitative examples from TextVQA, where the first 6 cases are successful and the last 2 are failures. Next to each image, we present the question alongside responses from GPT 4.1 nano, and GPT 4.1 nano paired with the \textbf{CropVLM} model that accepts images at 2048x2048 pixels of input resolution, and which was trained using log-likelihood rewards. The red bounding box denotes the \textbf{CropVLM} proposed region of interest.}
    \label{fig:textvqa-qual-examples}
\end{figure*}

Figure \ref{fig:v-star-qual-examples} presents qualitative examples, illustrating the performance of \textbf{CropVLM}, when paired with GPT 4.1 nano as the target model, on the out-of-domain V* benchmark. Results are presented for a \textbf{CropVLM} model trained using log-likelihood rewards, considering 2048x2048 pixels of input resolution. In turn, Figure \ref{fig:textvqa-qual-examples} presents qualitative examples for a similar setting, but instead considering instances from the TextVQA benchmark. Both cases present success cases, in which GPT 4.1 nano was originally unable to provide the correct answer but the inclusion of an image crop lead to a different result, and also failure cases in which the image crop provided by \textbf{CropVLM} distracted the model from providing the correct result.

\newpage

\end{document}